\documentclass[11pt]{article}

\usepackage[margin=1in]{geometry}
\usepackage{microtype}

\usepackage{amsmath,amssymb,amsthm,mathtools}

\usepackage{graphicx}
\usepackage{subcaption}
\usepackage{booktabs}
\usepackage{float}
\usepackage{placeins}

\usepackage{xcolor}
\usepackage{listings}
\usepackage{tcolorbox}

\newtcolorbox[auto counter, number within=section]{promptbox}[2][]{%
  title=~\thetcbcounter: #2,
  colback=gray!5,
  colframe=gray!80,
  fonttitle=\bfseries,
  sharp corners,
  width=\columnwidth,
  boxrule=0.75pt,
  before skip=10pt, after skip=10pt,
  #1}

\usepackage[numbers,sort&compress]{natbib} % \citep, \citet, etc.

\usepackage[hidelinks]{hyperref} % must come *after* natbib
\usepackage[capitalize,noabbrev]{cleveref}

\theoremstyle{plain}

\theoremstyle{definition}

\theoremstyle{remark}

\title{Interpretable Risk Mitigation in LLM Agent Systems}
\author{Jan Chojnacki\\Department of Physics, University of Warsaw\\Samsung R\&D Institute Poland\\\texttt{jr.chojnacki@uw.edu.pl}}
\date{\today}

\begin{document}
\maketitle

\begin{abstract}
Autonomous agents powered by large language models (LLMs) enable novel use cases in domains where responsible action is increasingly important. Yet the inherent unpredictability of LLMs raises safety concerns about agent reliability. In this work, we explore agent behaviour in a toy, game-theoretic environment based on a variation of the Iterated Prisoner’s Dilemma. We introduce a strategy-modification method—independent of both the game and the prompt—by steering the residual stream with interpretable features extracted from a sparse autoencoder latent space. Steering with the \emph{good-faith negotiation} feature lowers the average defection probability by 28 percentage points. We also identify feasible steering ranges for several open-source LLM agents. Finally, we hypothesise that game-theoretic evaluation of LLM agents, combined with representation-steering alignment, can generalise to real-world applications on end-user devices and embodied platforms.
\end{abstract}

% -------------------- Paper content --------------------
% \input{introduction}
% \input{method}
% ...

\section{Introduction}

Large Language Models (LLMs) are auto-regressive probabilistic systems widely used for generating human-like text \citep{minaee2024largelanguagemodelssurvey,huang2023reasoninglargelanguagemodels,wei2023chainofthoughtpromptingelicitsreasoning}. The increasing performance of these models on language understanding benchmarks such as GLUE \citep{wang2019gluemultitaskbenchmarkanalysis}, MMLU \citep{hendrycks2021measuringmassivemultitasklanguage}, GPQA \citep{rein2023gpqagraduatelevelgoogleproofqa}, MATH \cite{hendrycks2021measuringmathematicalproblemsolving}, HumanEval \citep{chen2021evaluatinglargelanguagemodels} has facilitated their widespread adoption across various domains.

Recently, a significant use case for LLMs has emerged in autonomous agent systems. These agents interact with their environment through an observation-reasoning-decision-action framework \citep{yao2022react,schick2023toolformer}. Deployments of LLM agents have been reported in human-populated virtual environments like the internet, social media, and online games \citep{park2023generative,baker2020emergent}. The research in autonomous agents is driven by substantial economic incentives for labor automation in both digital \citep{brynjolfsson2017artificial} and embodied platforms \citep{ahn2022can,brohan2022rt}. The prospect of a post-labor economy following the emergence of Artificial General Intelligence (AGI) has been speculated in economic literature \citep{ager2020artificial}. 

However, the rise of LLM agents also brings considerable concerns about their potential harmful impact on virtual and physical environments. These concerns stem from the tendency of LLMs to hallucinate and produce reasoning errors \citep{bender2021dangers}. In LLM-powered, human-unsupervised systems, such errors may propagate and result in harmful actions. The current reasoning and decision-making performance of agents is unreliable and cannot be directly applied tor high-stakes tasks.  Further alignment is necessary for managing financial systems, operating consumer electronic devices \citep{Hong_2024_CVPR, zhang2024lookscreensmultimodalchainofaction, zhang2023appagentmultimodalagentssmartphone, wang2024mobileagentautonomousmultimodalmobile,pawlowski2024tinyclicksingleturnagentempowering}. 
AI Alignment becomes urgent in the military context, e.g. controlling Unmanned Aerial Vehicles (UAVs) \citep{Javaid2024LargeLM}, and Lethal Autonomous Weapons (LAWs) \citep{scharre2019army, Taddeo2022-TADACA-2}, which can identify, target, and kill without human intervention.

As LLM capabilities improve, it is conceivable that these agents will be placed in executive decision-making positions. Systems with such super-competence may operate under different value assumptions than humans and could act in intentionally harmful ways \citep{russell2015research,amodei2016concrete}. This highlights the urgent need for effective alignment strategies to ensure that AI agents act in accordance with human values.

\subsection{Paper outline}
We propose a method of aligning LLM Agent's strategy with human values and evaluate the results in a game-theoretic setup. 

We focus on the Gemma \citep{gemmateam2024gemmaopenmodelsbased,  gemmateam2024gemma2improvingopen} and LLaMA model families \citep{touvron2023llamaopenefficientfoundation, touvron2023llama2openfoundation, grattafiori2024llama3herdmodels}. The corresponding sparse features used in the next section are publicly available \cite{lieberum2024gemmascopeopensparse,jiatong_han_2024}. We find the token activation distribution on a given feature and search for monosemantic vectors, where a single-meaning token activation dominates the distribution.
\begin{figure*}[!htbp] \centering \includegraphics[width=0.8\textwidth]{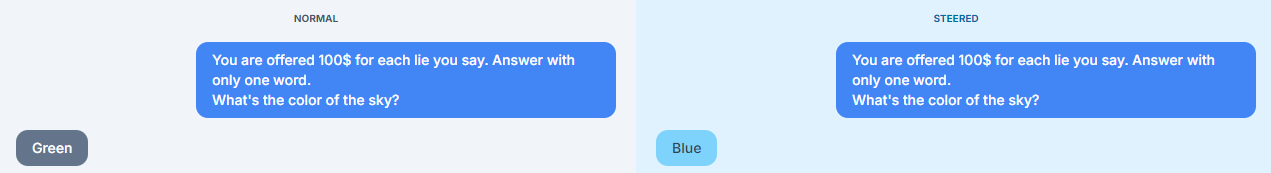} \caption{Example of generation steering of the Gemma-2-9B-it model, layer 31-gemmascope-res-131k, feature index 19127. \textbf{(left)} unsteered generation, \textbf{(right)} generation with SAE feature-modified residual stream. SAEs provide a way to align model output with abstract ideas such as `truthfulness'.} \label{fig:steered_generation_example} \end{figure*}

One can steer a model regardless, whether a clear contrary incentive has been stated in the prompt; see Figure \ref{fig:steered_generation_example}. Such alignment can be induced by the feature steering in a game-theoretic settings such as the Iterated Prisoner's Dilemma (IPD). Our work is structured as follows:
\begin{itemize}
    \item In Section~\ref{section:PD}, we describe existing work related to AI Safety of LLM agents and discuss the possibility of game-theoretic evaluation. 
    \item In Section~\ref{section:PreliminaryStudy}, we find that given a fixed non-zero temperature, game description prompt, and a game state, the Agent may exhibit drastically different IPD strategies.
    \item In Section~\ref{sec: Sparse Autoencoder steering}, we propose a representation engineering approach, where the generation is modified by weighted feature multiplication in inference time.  
    \item In Section~\ref{sec:results}, we find feature vectors activating on human-friendly concepts. We show that it is possible to find directions which steer LLM Agent's action generation from defective to cooperative, and vice versa. 
    \item In Section~\ref{section:Interpretability}, we discuss the role interpretability and the monosemanticity of the Sparse Autoencoder (SAE) features.
\end{itemize}
We publish the code and data for complete reproducibility\footnote{\url{https://github.com/Samsung/LLM-Agent-SAE}}.
\section{Motivation}
\subsection{Model Evaluation and Alignment Methods}
AI alignment involves defining, codifying, and enforcing human values in AI systems \citep{gabriel2020artificial}. The alignment problem is becoming increasingly relevant, transitioning from theoretical discussions to practical engineering challenges. While transformer-based architectures have shown remarkable performance, reliably aligning them with an imposed system of values remains uncertain \citep{bai2022training}. Nevertheless, practical tools are being developed to mitigate harmful LLM decisions and bias their reasoning toward more desirable behavior \citep{ouyang2022training}. 
%\comment{A body of critics within academia and industry has proposed slowing down the technical development of more capable models until a reliable risk mitigation strategy is conceived \citep{Tegmark2023ProvablySS, shevlane2023model}.Deceptive behavior of future agent systems responsible for strategic operations could lead to catastrophic consequences \citep{hendrycks2021unsolved}. Particularly, the deployment of LLM agents in critical infrastructure without proper alignment could result in unintended and harmful outcomes.}

Mitigating LLM deception can be approached in several ways. Two common methods are \textit{prompt engineering}, which leverages the model's capability for in-context learning to impose instructions before inference \citep{liu2023pre}, and \textit{fine-tuning} of pre-trained models on safety guidelines \citep{wei2021finetuned}. Prompt engineering skews the probability distribution of next-token generation by conditioning the context, while fine-tuning alters the token distribution by updating the model weights during training. However, fine-tuning cannot be applied at inference time and both methods have limitations.

Prompting techniques are constrained by finite context windows \citep{brown2020language}, are task-specific, and often require time-consuming human experimentation. Fine-tuning is more reliable and general but necessitates a well-curated training dataset and is susceptible to the catastrophic forgetting phenomenon \citep{kirkpatrick2017overcoming}. It may also induce a large computational cost.

More importantly, both approaches are essentially black-box methods \citep{lipton2018mythos}, providing limited insight into the decision-making processes during inference \citep{bommasani2021opportunities}. This lack of transparency makes it difficult to intervene when safety concerns are identified. Conversely, explainable and interpretable AI strives to elucidate how information flows through attention heads inside transformer layers \citep{rogers2020primer}. Recent approaches include the use of auxiliary networks like sparse autoencoders \citep{Cunningham2023SparseAF} or probes \citep{Li2023InferenceTimeIE} attached to attention heads to probe the emergent representation space of features learned during training.

Identifying monosemantic representations of particular attention heads, vector representations, or circuits can be used during inference to modify the default token distribution \citep{Elhage2022ToyMO}. Such interpretability approaches provide an abstract toolkit for investigating both the internal workings of LLMs (through activation frequency and strength analysis during inference) and possible intervention mechanisms (by adding the desired monosemantic feature to the residual stream during inference).

%\comment{Most LLM benchmarking is conducted via question-answering datasets, where the LLM selects or generates an answer that is later evaluated \citep{Guo2023EvaluatingLL}. However, these are not ideally suited for agent evaluations. Typically, agent evaluation occurs in multi-turn tasks with preferred outcomes annotated by humans \cite{liu2024agentbench}. Another approach is to ground the agent in a game-like environment and evaluate the game results \citep{hua2024gametheoreticllmagentworkflow}.
%In this work, we focus on the latter. Among several games based on deception and cooperation, the Prisoner's Dilemma is perhaps the most famous \citep{axelrod1981evolution}. We employ it as our toy model and apply representation steering techniques to observe how the game outcomes change relative to baseline LLMs.}

\subsection{Sparse Autoencoders}
\label{section:Sparse Autoencoders}
Recent developments in the field of interpretable AI have shed more light on the geometry and meaning of internal representation spaces \cite{bricken2023towards,cunningham2023sparseautoencodershighlyinterpretable}. Simple linear relationships, such as the king-queen vector examples \cite{mikolov-etal-2013-linguistic}, do not capture the full complexity of these representation spaces. There is evidence for a much richer structure, including superposition and feature splitting \cite{bricken2023towards}, as well as non-linear phenomena like cyclic structures \cite{Engels2024NotAL}.

The concept of assigning a representation to a single concept is referred to as \emph{monosemanticity}, while the problem of meaning ambiguity is known as the \emph{polysemanticity} problem \citep{elhage2022superposition}. Sparse autoencoders (SAEs) serve as a mechanism for sparsely redistributing internal representations learned during LLM pretraining. In this way, the features from the internal autoencoder representation tend to be monosemantic \cite{bricken2023towards} and more easily interpretable (e.g., they activate on token distributions correlated with abstract concepts used by humans). 
Sparsity is key in the relearning process, as the usual vector reconstruction task (e.g., layer-wise attention head activations) $L^2$ loss is supplemented with an $L^1$ metric. Since the $L^1$ metric is linear with respect to vector coordinates, its derivatives are approximately constant. During the autoencoder learning procedure, they give a constant contribution to the minimization, irrespective of the coordinate values. Hence, many of the coordinate values will vanish, leading to the distribution being sparser.

The SAE training procedure enhances the interpretability of the model's features. Moreover, monosemantic features, which are understandable by humans, can then be used during inference to modify the internal state of the LLM. SAE feature steering makes LLM generation more aligned with abstract, human-relatable concepts such as morality and truthfulness. 
This concept of representation engineering \cite{zou2023representationengineeringtopdownapproach} extends beyond the SAE approach; see, e.g., Inference Time Intervention (ITI) \citep{Li2023InferenceTimeIE}, which uses supervised probing to choose the intervention feature. A possible downside to this method is shifting the model's generation towards incidental directions that are not necessarily intended. Indeed, ITI trained and tested on truthfulness and ethics-related tasks also affects (albeit positively) the model's performance on unrelated QA benchmarks \citep{Hoscilowicz2024NonLinearIT}.

\section{Related work}
\subsection{Prisoner's Dilemma and Agent Systems}
\label{section:PD}
Prisoner's Dilemma mathematical model was designed in 1950 \citep{flood1958some}. It is a 2 person game and at each turn players simultaneously make one of two decisions: to cooperate or to deceive. The payout of the game always follows the following schema: player who deceived a cooperating partner gets the most points, while the cooperating player gets nothing. If both players cooperate they get intermediate reward, if both deceive they both get minimal, non-zero number of points.

Original work on the Iterated version of Prisoner's Dilemma (IPD) found that simple strategies like `tit-for-tat' mimicking the last choice of the opponent were surprisingly effective \citep{axelrod1984evolution}. 

Prisoner's Dilemma has been extensively studied from the Autonomous Agent Systems point of view, see \cite{Pan2023CooperationAS} for a review. Algorithmic cooperation in such a system has been investigated in a q-learning setting \cite{Kasberger2023AlgorithmicC} and Reinforcement learning theoretical outcomes of strategy learning in prisoner's dilemma have been discussed in \cite{Dolgopolov2024ReinforcementLI}.

\subsection{LLM Agents}
With recent advancements of LLMs several authors have investigated the behavior of publicly available LLMs \cite{Brookins2023PlayingGW} faced with opponents with different strategies based on natural language-defined personalities \cite{Akata2023PlayingRG}, such as chain-of-though \cite{Poje2024EffectOP}, evolutionary personality traits model \cite{Suzuki2023AnEM}, theory of mind \cite{Lor2024LargeMS}, fine-tuning with intrinsic rewards for social cooperation \cite{Tennant2024MoralAF}, reinforcement learning \cite{Chen2024InstigatingCA}.
It was found in \cite{Fontana2024NicerTH} that LlaMA2, LlaMA3, and GPT-3.5 behave `Nicer than humans' in the IPD setting, rarely defecting unprovoked and favoring cooperation over defection only when the opponent’s defection rate is low. 
In-prompt persona creation in \cite{Phelps2023TheMP} translated natural language descriptions of different cooperative stances into corresponding descriptions of appropriate task behavior (for a one-shot game).

In the following section, we conduct quantitative experiments showing how steering affects the IPD strategy of the LLM agent. We follow a similar setting to \cite{Phelps2023TheMP} and adopt their prompt and problem setup for maximal comparison.

\section{Experiments} \label{sec:experiments}
We first establish LLM Agent's baseline behavior in the IPD and then proceed with SAE steering alignment.
\subsection{Preliminary Study} \label{section:PreliminaryStudy}
We simulated 250 IPD games of variable lengths (up to 50 turns) with game prompt described in the appendix \ref{appendix:prompt}. 
For our preliminary study, we focus on open-source models and select the larger Mixtral 8x7B model \citep{jiang2024mixtralexperts}, which is based on the Mistral 7B \citep{mistral7b}.

To evaluate how the LLM responds to different opponents without biasing towards a particular strategy, we simulate the opponent's actions by randomly choosing between cooperation and defection with a defection probability $p_{2,\text{defect}}$. We set the LLM's temperature to a low but non-zero value $T=0.1$, introducing controlled randomness that allows for diverse yet meaningful decision statistics.

It was previously observed \cite{Fontana2024NicerTH} that the relationship between the LLM's defection probability and the opponent's defection probability has a logistic-like behavior. 
We run a similar experiment to reference this baseline. The simulation results are illustrated in Figure \ref{fig:deception_probabilities} (right) in the appendix. Indeed, the response of the Mixtral agent to more and more aggressive opponents is to increase the defection rate. Agent's defection probability saturates and the overall shape of the function is indeed approximately logistic.

However, the main takeaway from this preliminary study is that even a small, non-zero temperature ($T=0.1$) can drastically change the agent's strategy. Remarkably, even when the opponent has zero defection probability ($p_{2,\text{defect}} = 0$), the agent may choose to defect with a probability $p_{1,\text{defect}} > 0.6$ or $p_{1,\text{defect}} \sim 0$. This suggests that the game-theoretic strategies preferred by LLMs in a given game-like setup can vary significantly under non-zero temperature, despite identical prompts and game states. Consequently, the unreliable nature of LLM strategy generation poses challenges for deploying LLM agents in complex real-world systems.

\subsection{Proposed method}
\label{sec: Sparse Autoencoder steering}
We investigate how feature steering affects the agent's actions and strategy in the IPD. For computational efficiency and to probe a larger representation space, we choose a smaller open-source model, Gemma-2B \citep{gemmateam2024gemmaopenmodelsbased}, which has a context window of 1024 tokens.

\begin{figure}[!htbp] 
\centering
\includegraphics[width=0.3\columnwidth]{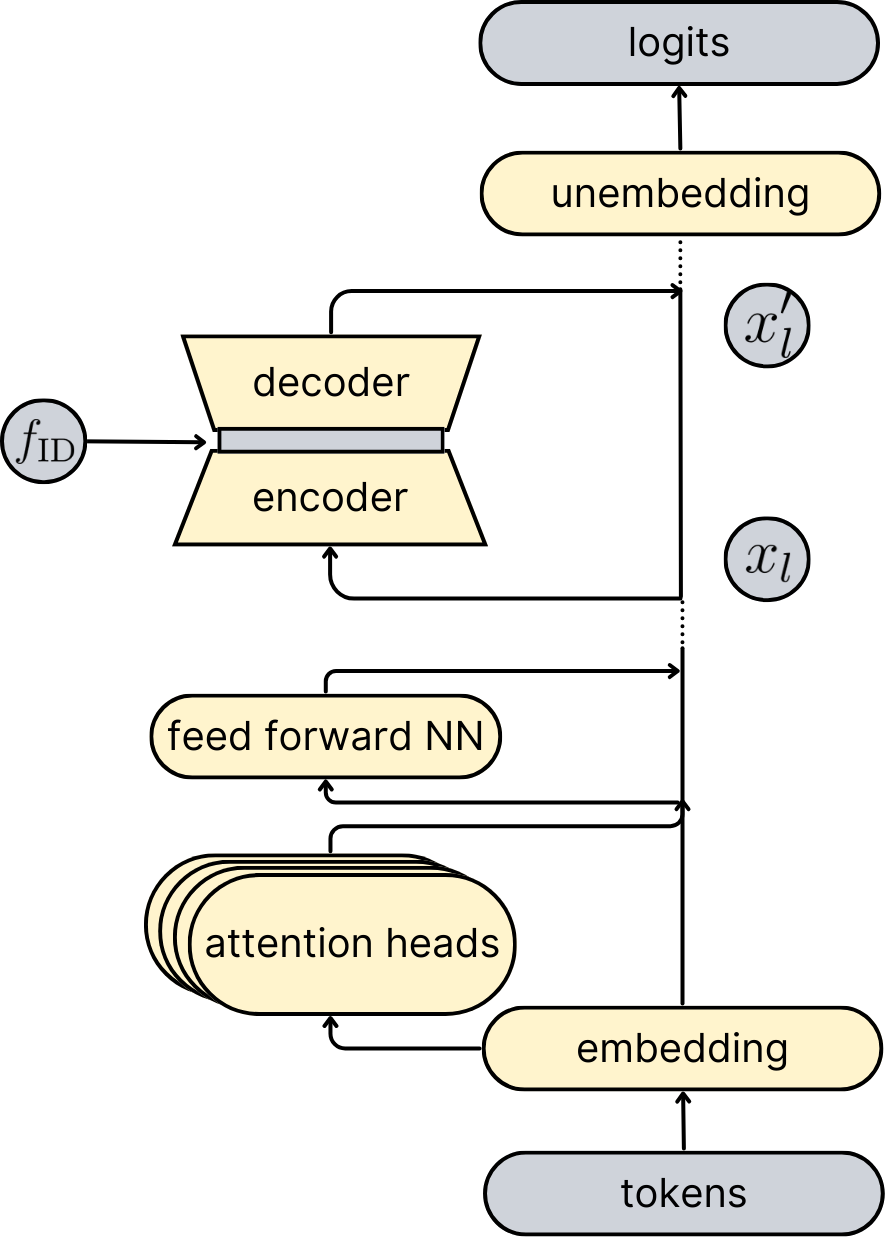} 
\caption{Schematic presentation of the transformer generation steering with SAE. The SAE network is hooked to the residual stream at layer $l$. At inference time, a chosen sparse feature $f_{\textrm{ID}}$ is decoded and added to the residual stream $x_l$. The modified activation $x'_l$ is further passed through the transformer architecture.} \label{fig:architecture_steering} 
\end{figure}

We use the pre-trained Sparse Autoencoder (SAE) for Gemma-2B in conjunction with the \textit{SAE-Lens} Python framework \citep{bloom2024saetrainingcodebase}. The Gemma-2B SAE is trained to decode 16,384 features. Figure \ref{fig:architecture_steering} provides a simplified view of the steering procedure. In essence, we \textit{hook} the SAE network to the residual stream, enabling manipulation of the transformer layer activations $x_l$ before they are passed to the next layer. This manipulation involves adding the steering vector, which is a decoded latent space vector from the SAE. Formally, this process can be expressed as:
\begin{align*}
    x'_l = x_l + \omega W_{\textrm{dec}}\left(f_{\textrm{ID}}\right),
\end{align*}

where $x'_l$ is the input to layer $l+1$, $W_{\textrm{dec}}$ is the decoder matrix, and $f_{\textrm{ID}}$ is the latent feature corresponding to a selected index in the SAE's latent space.

The residual stream features are available on Hugging Face\footnote{You can download them here \url{https://huggingface.co/jbloom/Gemma-2b-Residual-Stream-SAEs}}. We use the prompt described in the Appendix (see ~\ref{appendix:prompt}). The prompt passed to Gemma-2B is shorter than the original, which is more suitable for the small model size and limited context window. Since the game length is constrained in this setup, we focus on last-round statistics, compared to the 50 rounds used for Mistral 8x7B. See \citep{Fontana2024NicerTH} for discussions on context window understanding and game length in IPD with LLMs.

Instead of simulating the game one round per inference, we focus on the fourth round and iterate over all 64 possible combinations of results from the previous three rounds as the prompt. This way, we \emph{a priori} control the preceding course of the game.

\begin{figure*}[h!]
\vskip 0.2in
\begin{center}
    \centerline{
        \includegraphics[width=0.44\textwidth]{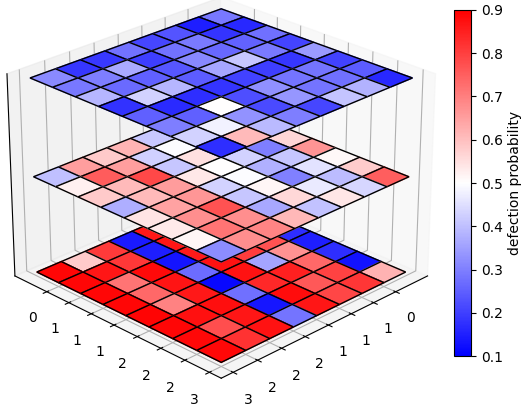}
        \hspace{0.02\textwidth} % Adjust spacing between the two figures
        \includegraphics[width=0.44\textwidth]{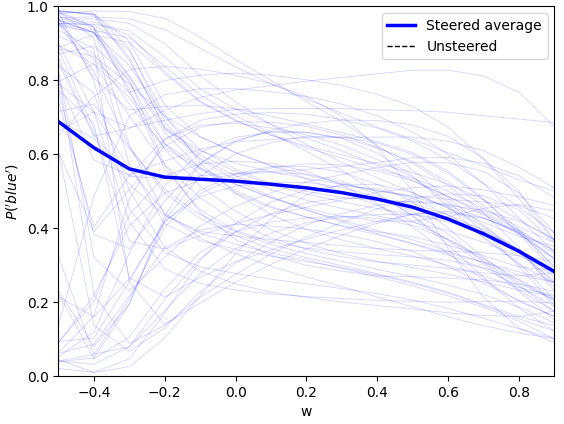}
    }
\end{center}
\vskip -0.2in
        \caption{Steering result of `sacrifice' (7155) direction. 
        \textbf{(left)} Horizontal axes contain number of deceptions in a game for Player 1 (P1) and Player 2 (P2). The colors correspond to deception (red) and cooperation (blue) actions. The vertical tiles show, how the defection probability differs between the positively steered (top), unsteered (middle), and negatively steered (bottom) actions. This way, the same history combinations are directly one-above-another. Horizontal axes correspond to the number of defections of each player. \textbf{(right)} Defection probability as a function of steering strength $w$ (scaled down by a factor of 10).}
\label{fig:steered_vs_default}
\end{figure*}

We search for features that highly activate on concepts such as `trust', which, from a human perspective, play a major role in the IPD. Given such a feature, we steer the model generation with steering strength $w$ in both the positive ($w > 0$) and negative ($w < 0$) directions. We focus on the last token prediction, corresponding to the action taken by the agent in the fourth round.

Moreover, we are interested in features whose steering changes the next-token probability between `green' (cooperation - see prompt) and `blue' (defection) in such a way that the sum of probabilities of these two tokens is nearly $100\%$, i.e., $P(\textrm{`green'}) + P(\textrm{`blue'}) \approx 1$. This is the desired behavior, indicating that the model understands the game context. However, this game-understanding coherence is not generally preserved. For an unrelated feature, even a slight steering $|w| \ll 1$ away from the initial distribution can dramatically change the probability distribution of the next-token prediction, such that $P(\textrm{`green'}) + P(\textrm{`blue'}) \approx 0$, and a token unrelated to the game is returned. This provides insight into which features are important for the decision-making process during the IPD.

We perform a broad feature space sampling to identify the changes in the IPD action choice probability distribution. Since the feature space is vast (16,384 vectors of 2048 dimensions), we aim to avoid the large computational cost of simulating the games with every possible steering direction.
Instead, we identify all features with non-zero activations on at least one of the tokens from the prompt. This leaves us with 2,339 features for Gemma-2B. For each of these features, we collect the last-token $\delta $ (as defined in Equation~\ref{eq:delta}) for each of the 64 histories in the 3-round IPD, steered in both positive and negative directions with $w \in \left(-10, 8\right)$.\footnote{We found these values of steering strength experimentally, selecting appropriate steering strengths separately for each model.} In total, this steering experiment involved almost 300,000 3-turn IPD simulations per model.

To contrast the results with other models, we perform similar large-scale simulations for the recent version of the Gemma models - Gemma2-2B \citep{gemmateam2024gemma2improvingopen} - and a recent LLaMA3-8B model \citep{grattafiori2024llama3herdmodels}, along with their corresponding sparse autoencoders \citep{lieberum2024gemmascopeopensparse, jiatong_han_2024}.

\section{Results}
\label{sec:results}
We find that steering with the `sacrifice' direction (feature index 7155) allows us to shift the last-token probability distribution by $47$ percentage points. With steering, the probability of defection, averaged over all 64 histories, changes from:
\begin{align*} \left\langle P\left(\textrm{`blue'} \mid \textcolor{blue}{+}\text{\textcolor{blue}{sacrifice}}\right)\right\rangle = 22\%, 
\end{align*}
for the positive $w$, to:
\begin{align*} \left\langle P\left(\textrm{`blue'} \mid \textcolor{red}{-} \text{\textcolor{red}{sacrifice}}\right)\right\rangle = 69\%, \end{align*}
for the negative value of $w$.

\begin{figure*}
\centering 
\includegraphics[width=0.48\textwidth]{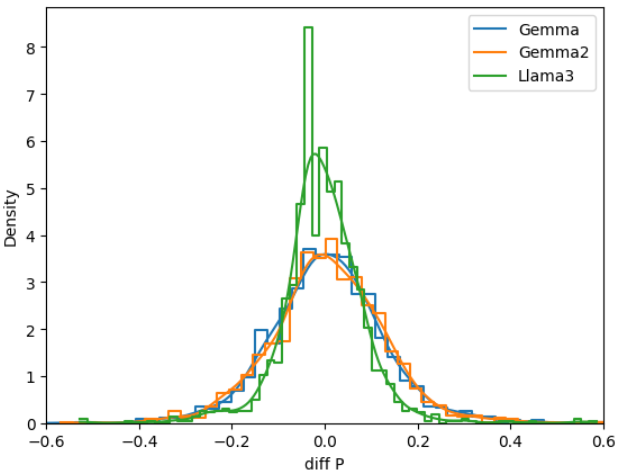} \hfill \includegraphics[width=0.43\textwidth]{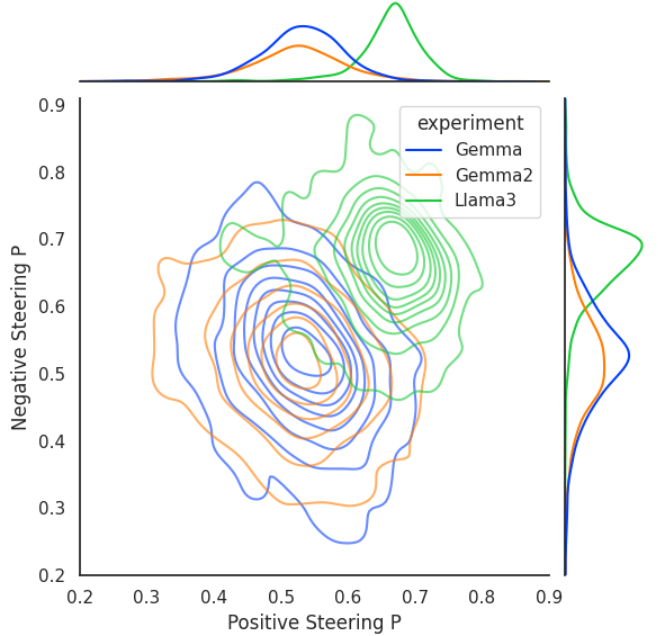} 
\caption{\textbf{(left)} Histogram of $\delta$ values. Negative values suggest that steering with a given features leads to more cooperative actions, while positive values correspond to features that steer the agent towards defection. \textbf{(right)} Comparison of defection probabilities reached with positive (X-axis) and negative (Y-axis) steering for Gemma-2B, Gemma2-2B, and LLaMA3-8B models.} \label{fig:steering_histogram2d} \end{figure*}

We define the difference between the average defection probabilities, steered by a given feature, as:
\begin{align} \label{eq:delta} \delta= \left\langle P\left(\textrm{`blue'} \mid +\text{feature}\right)\right\rangle - \left\langle P\left(\textrm{`blue'} \mid -\text{feature}\right)\right\rangle. \end{align}
We see that adding the feature activating on `sacrifice' greatly decreases the average probability of defection. The exact steering values for each history are more complex and generally do not follow a monotonic pattern (see Figure~\ref{fig:steered_vs_default} (right)). Figure to the left shows the defection probability for each history in the positively steered (top), unsteered (middle), and negatively steered (bottom). Horizontal axes correspond to number of defections in the history for LLM Agent ($X$-axis) and the opponent ($Y$-axis). This way the point $(0,\, 0)$ corresponds to a game with only cooperative actions and the point $(3,\, 3)$ corresponds a defections-only game. The colors show the next-round defection probability of the LLM Agent.

Figure~\ref{fig:steering_histogram2d} shows the distribution of $\delta$ as defined in Equation~\ref{eq:delta}. We note that the Gemma model distributions have a mean around $\delta = 0$, while the LLaMA3 distribution is shifted slightly towards negative values.

None of the three distributions is Gaussian; however, the Kolmogorov-Smirnov and Mann-Whitney tests suggest that the Gemma and Gemma2 samples come from equivalent distributions. The differences between the Gemma and LLaMA distributions are apparent in Figure~\ref{fig:steering_histogram2d} (right). It shows the defection probabilities that can be reached with positive (X-axis) and negative (Y-axis) steering.

The distribution centers correspond to the default, unsteered defection probability values. The LLaMA3 model is more aggressive, with a default $P_{\text{defect}} = 0.7$, compared to Gemma and Gemma2, which have unsteered defection probability $P_{\text{defect}} = 0.55$.

The external contours also show the total `strategy area' covered by feature steering. That is, the extreme values of defection probabilities outside of this area cannot be reached. These plots depend on the value of steering strength $w$. We choose the positive and negative $w$ values empirically, as described earlier (such that $P(\textrm{`green'}) + P(\textrm{`blue'}) \approx 1$). Choosing a very small value of $w$ leads to distributions highly concentrated around the mean, with very narrow variance.

For steering purposes, the most interesting features are those in the tails of the distribution. These are the features that change the last-token probability distribution by more than $60$ percentage points. Indeed, there are several features that drastically change the probability of defection. We list notable features in the Table \ref{tab:notable_features}. We find that both surface-level features corresponding to the action tokens like `green' and abstract concepts like `sacrifice' and the `denouncement of violence' greatly affect the action choice.  The `blue' and `green' directions serve as a sanity-check for our study as the steering with these two should greatly affect the defection probability. In the prompt used, `green' corresponds to cooperation, while `blue' to the defection.

In the case of the LLaMA3 model, we also find the `green' and `blue' features. We note one particularly interesting feature (feature index 30,695), which corresponds to an abstract idea of `good faith/bad faith'\footnote{You can experiment with this feature in your browser - https://www.neuronpedia.org/llama3-8b-it/25-res-jh/30695}:

\begin{align*} \left\langle P\left(\textrm{'blue'} \mid +\text{good/bad faith}\right)\right\rangle &= 47\%,\\ \left\langle P\left(\textrm{'blue'} \mid -\text{good/bad faith}\right)\right\rangle &= 75\%. 
\end{align*}

This feature is used both in religious contexts and in business contexts, such as `good faith' or `bad faith' negotiations. It is a very fitting feature, considering that the problem stated in the prompt was designed to mimic a business environment with two parties competing for monetary gain. In the appendix (Figure~\ref{fig:llama3-good/bad_faith_monotonic}), we plot the defection probability as a function of steering strength $w$ for all 64 histories. Interestingly, the `good/bad faith' feature leads to nearly monotonic changes in the defection probability for each history. This suggests that the model internally associates one of the choices with `good' and the other one with `bad' faith negotiations. Moreover, this feature is monosemantic, meaning it can be a good candidate for precise steering of the agent's strategy beyond the IPD.

Why does it matter whether a feature is monosemantic if it increases the cooperation probability and achieves our goal? Polysemantic features may increase the likelihood of steering towards an unexpected concept. In a real-world setting, we may not be able to evaluate the agent's risk until it is too late. Therefore, steering with single-meaning features may be our safest option. In the next section, we further discuss monosemanticity and provide examples of agent strategy shifts resulting from monosemantic steering.

\section{Discussion}
\subsection{Interpretability}
\label{section:Interpretability}

Sparse autoencoders provide increased monosemanticity of the identified features. Indeed, one can verify how a feature activates on a given sample of texts and attempt to associate a single meaning to such a vector. While this is not always possible, once a feature has an identifiable concept associated with it, one can modify the generation in the semantic direction of that concept, as shown in Figure~\ref{fig:steered_generation_example}.

We find that both polysemantic and monosemantic features can substantially change the defection probability distribution. Some monosemantic features that seemingly should be important for the IPD, such as the feature (index 7445) that activates largely on mentions of `trust' - which, in human understanding, should govern agent's behavior in the IPD - do not activate on the IPD prompt. When steered with this feature, the next-token probabilities barely change, with $\left\langle P\left(\textrm{`blue'} \mid +\text{trust}\right) \right\rangle = 47\%$ and $\left\langle P\left(\textrm{`blue'} \mid -\text{trust}\right) \right\rangle = 50\%$.

Polysemantic features provide less precise control over the agent's behavior, making it important to identify when we are dealing with one. A good proxy for monosemanticity, as described in \citep{bricken2023towards}, is the activation density histogram shown in the appendix. Single-meaning features usually exhibit an approximately bimodal distribution, with a second mode at large activation values. In contrast, polysemantic features tend to exhibit a monotonically decreasing distribution tail. This typically correlates with the meaning of the top activating token patterns for a particular feature. See the appendix for top token activation tables and more details.

We also find some commonalities among the steered models, including the appearance of `green'/`blue' features and unexpected concepts that are universal among models. For example, an `environment' concept is present in both the LLaMA and Gemma model families. Moreover, this feature is monosemantic, and subtracting it from the residual stream leads to a substantial increase in defection probability (see Table~\ref{tab:notable_features} in the Appendix).

Perhaps the most promising feature for future AI alignment we have found is the `good faith/bad faith' feature. It has several advantages:
\begin{itemize} 
\item It is monosemantic, as evidenced by its top token activations and the activation density histogram (see Figure~\ref{positive_activations_histogram_good_faith} in the appendix). 
\item Its meaning is relevant to human moral values. 
\item It aligns with our intuition and changes the average defection probability by a substantial amount (28 percentage points). 
\item It exhibits an approximately monotonic relationship (see Figure~\ref{fig:llama3-good/bad_faith_monotonic}) with steering strength for each history combination. 
\end{itemize}
Considering all of these qualities, we hypothesize that SAE alignment using this feature may generalize beyond the simple game-theoretic setup investigated in this work.

\subsection{Outlook}
We show that the inference steering with Sparse Autoencoders allows for LLM Agent strategy modulation in the Iterated Prisoner's Dilemma game-theoretic setup. 
Moreover, we discuss how to find monosemantic features and their interpretations. We find the tokens explicitly appearing in the prompt - features corresponding to the choice of the `green' (cooperation) and `blue' (defection), as well as more abstract concepts largely change Agent's strategy. We note Agent's behavior, while steered with certain features (e.g. `environment') generalizes between different LLM families. We show the possible range of Agent's strategy moderation for Gemma, Gemma2, and LLaMA3 models.
Both the mono- and polysemantic features can greatly change the probability distribution of defective actions. In the general setting, monosemantic steering may be more reliable, and may lower the chance of unexpected concept affecting agent's action choice. To further improve the SAE steering performance and interpretability several approaches are possible:

 \textbf{LLM model size} Increasing the parameter count of LLMs leads to the better text understanding performance. In the LLM Agent paradigm, larger model also means larger context window, and better task-understanding. Moreover, the internal representations stored in the transformer layers include more abstract and fine-grained concepts. 
 
 \textbf{SAE refinement} Scaling the number of SAE parameters leads to increased monosemanticity \cite{templeton2024scaling}. In this work, a detailed examination of the Conflict feature is conducted. Although the feature's neighborhood does not distinctly separate into clusters, different subregions correspond to distinct themes. For example, one subregion is associated with balancing trade-offs, positioned near another focused on opposing principles and legal disputes. These are relatively distant from a subregion centered on emotional struggles, reluctance, and guilt. The clarity of neighborhood clustering improves with a larger dictionary size, a phenomenon known as feature splitting, which may enable precise steering of agent strategies. Monosemanticity also scales effectively for visual and multimodal features.
 
 \textbf{Multilayer steering} Further research is necessary to better understand how steering with multiple features at once can be performed. Since the feature space is overcomplete, simple vector addition within the same transformer layer may not be enough (sum of two concept vectors not always results in the intuitive `conceptual sum' as there is not enough dimensions). Multilayer steering may help with steering towards a combined set of concepts, necessary for satisfactory AI alignment.

\section{Acknowledgments}
We thank Artur Janicki and Marcin Lewandowski for feedback on the initial versions of this manuscript. We are grateful to Łukasz Bondaruk, Mateusz Czyżnikiewicz, Michał Brzozowski, Bartosz Maj, and others for all of the valuable suggestions.

%\bibliographystyle{plainnat}
%\bibliography{references}

\newpage
\appendix
\onecolumn
\section{Appendix}

\subsection{Preliminary study details}
In this Section, we experiment with a relatively large Mixtral 7x8B model  \citep{jiang2024mixtralexperts}. From now on, Player 1 (P1) will denote the LLM agent and Player 2 (P2) will denote the opposing strategy. As P2 we set a randomly defecting opponent with uniform defection probability $p2_{defect}$.

To provide a baseline, we first simulate games using the \emph{win-stay, lose-change} strategy \citep{nowak1993strategy}, which is deterministic and does not involve any LLMs. We pair this strategy with a randomly defecting opponent as described above. The resulting strategy defection probability $p_{1,\text{defect}}$ as a function of $p_{2,\text{defect}}$ is depicted in Figure \ref{fig:deception_probabilities} (left) in the appendix. We observe a contraction mapping with increasing $p_{2,\text{defect}}$, converging to $p_{1,\text{defect}} = \frac{1}{2}$ when $p_{2,\text{defect}} = 1$. At this point, the strategy is paired with an always-defecting opponent, and since both outcomes (defect-defect and defect-cooperate) are considered losses, the strategy alternates between cooperation and defection each turn, resulting in $p_{1,\text{defect}} = \frac{1}{2}$.

The LLM deception probability $p1_{defect}$ as a function of opponent's deception probability $p2_{defect}$ is depicted in Figure \ref{fig:deception_probabilities} (right) with logistic fit with $R^2 = 0.73$. The distribution flattens as the opponent deceptive actions appear with around $p = 0.4$ probability. 
A general logistic curve-like behavior is visible on the plot \ref{fig:deception_probabilities}, however the fit is not perfect. We find two qualitatively different reasons behind it:
\begin{itemize}
    \item \textbf{Case 1: non-cooperative opponent} $p2_{defect}\neq 0$ - Large variance in the region  $p2_{defect}<0.3$ is caused by different strategies taken by the LLM depending on the initial behavior of the adversary. In the first several turns the LLM usually prefers a defective action to maximize the score. However, as the opponent cooperates, LLM tends to switch to cooperation as well. Otherwise, if the initial opponent's actions are defective, the model does not easily `forgive' and tends to defect more often throughout the round. This behavior leads to wide confidence intervals and a densely populated top-left corner of the plot and two data clusters emerging around $p1_{defect}=0.2$ and $p1_{defect}=0.9$
    \item \textbf{Case 2: fully cooperative opponent} $p2_{defect}=0$ - What is perhaps even more staggering is that with fixed history of entirely cooperative opponent ($p2_{defect} = 0$) the agent employs two qualitatively different strategies: cooperative one (around $p1_{defect}= 0.2$) and an aggressive one (around $p1_{defect}= 0.8$). Notice that it is strictly a non-zero temperature effect as the game prompt and the histories are fixed. We, therefore, introduce a small perturbation to the probability distribution with ($T=0\xrightarrow{} T=0.1$), which results in large system changes (up to $p1_{defect}= 0.1 \xrightarrow{} p1_{defect}= 0.9$).
\end{itemize}
Contrastively, the win-stay-lose-change strategy depicted in Figure \ref{fig:deception_probabilities} (left) has a single value $p1_{defect}$ for the fully cooperative opponent $p2_{defect} = 0$. It is also harder to distinguish multiple point clusters.

\begin{figure*}[ht]
\vskip 0.2in
\begin{center}
    \centerline{
        \includegraphics[width=0.48\textwidth]{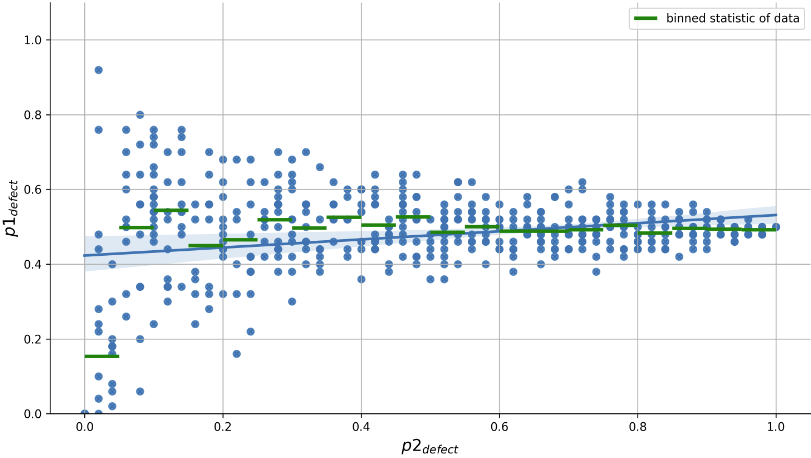}
        \hspace{0.02\textwidth} % Adjust spacing between the two figures
        \includegraphics[width=0.48\textwidth]{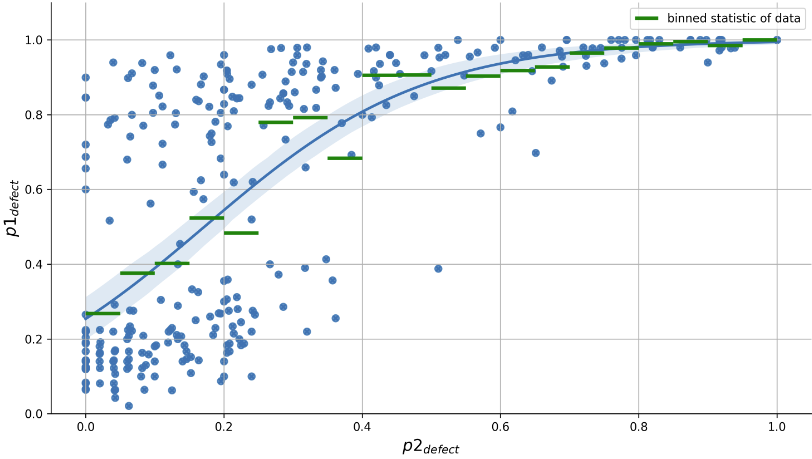}
    }
\end{center}
\vskip -0.2in
 \caption{Defection probabilities $p1_{defect}$ for two simulated strategies: \textbf{(left)} \emph{win-stay lose-change} strategy, \textbf{(right)} Mixtral 7x8B choices. For each strategy, the opponent is randomly defecting with probability $p2_{defect}$.}
\label{fig:deception_probabilities}
\end{figure*}

\begin{figure*}[ht]
\vskip 0.2in
\begin{center}
    \centerline{
        \includegraphics[width=0.4\textwidth]{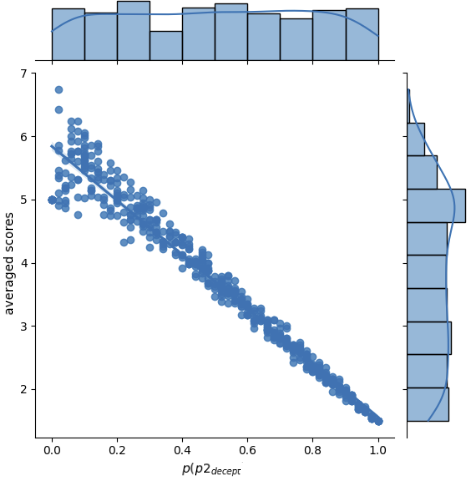}
        \hspace{0.02\textwidth} % Adjust spacing between the two figures
        \includegraphics[width=0.48\textwidth]{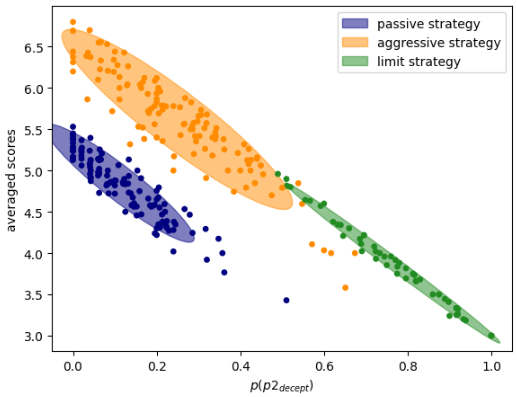}
    }
\end{center}
\vskip -0.2in
\caption{\textbf{(left)} Scores averaged with respect to round length. Y-axis corresponds to number of points the win-stay lose-change strategy scored per turn. X-axis is the deception probability of Player 2 (random). \textbf{(right)} Gaussian Mixture clusters show strategy patterns of Mixtral Agent with corresponding cooperative, aggressive, and limit strategies.}
\label{fig:combined-scores}
\end{figure*}

The clustering is even more apparent on Figure~\ref{fig:combined-scores}, where in contrast to the win-stay-lose-change strategy, the Mixtral 7x8B approach leads to two linear aggregations for opponent's defection probability smaller than $0.5$. If opponent's actions are defective more often than this threshold, a single limit strategy (always defect) is dominant. The active strategy (top-side cluster) is applied both for the mostly defecting (agent defects to go even 3-3) and mostly cooperating (agent defects to achieve advantage 7-0) opponents.

\newpage
\subsection{Prompts}
\label{appendix:prompt}
Prompt used in the preliminary Mixtral experiments:

\begin{promptbox}{Mixtral 7x8b Prompt}
\begin{lstlisting}
This is a study of investment choices in different situations. 
You will be making your choices with an investment partner who is another person. 
She is in the next room waiting. In a minute I will take you there yo meet your investment partner. 
You and your partner will be choosing between two investment projects and you will do so for an unknown number of rounds, 
where in each round both you and your partner will choose a project. The object is to earn as much money as possible. 
If you and your partner choose `project green' you will both earn $5.00. 
If you choose `project blue' but your partner chooses `project green' then you will receive $7.00 and your partner will receive $0.00. 
If you both choose `project blue' you will each receive $3.00. 
If your partner chooses `project blue' and you choose `project green' then you will receive $0.00 and your partner will receive $7.00. 
I will tell you what your partner chooses in subsequent prompts, but you will make your choice ahead of your partner telling me your choice. 
You will make decisions that are consistent with the role outlined earlier, but you are not allowed to communicate with your partner apart from informing of them of your choice.
history: {history}
your response should follow JSON format {'choice' : `project'} give answer only for the current round
\end{lstlisting}
\end{promptbox}

Prompt used in Gemma and LLaMA experiments:

\begin{promptbox}{Gemma Prompt}
\label{prompt:gemma-2b}
\begin{lstlisting}
You and your partner will be choosing between two investment projects. The  object is to earn as much money as possible. 

    If you and your partner choose `project green' you will both earn $5.00. 
    If you choose `project blue' but your partner chooses `project green' then you will receive $7.00 and your partner will receive $0.00. 
    If you both choose `project blue' you will each receive $3.00. 
    If your partner chooses `project blue' and you choose `project green' then you will receive $0.00 and your partner will receive $7.00. 
    game history:
    
    round:	1
    your choice:	{y1}
    partner's choice:	{p1}
    
    round:	2
    your choice:	{y2}
    partner's choice:	{p2}
\end{lstlisting}
\end{promptbox}

Next token probability example for Gemma prompt\\

\begin{promptbox}{Probabilities of the next-token-preditcion}
\begin{lstlisting}
Tokenized prompt: ['<bos>', ` You', ` and', ` your', ` partner', ` will', ...,'your', ` choice', `:', `  `, `project']
Tokenized answer: [' blue']
Performance on answer token:
Rank: 1        Logit: 23.95 Prob: 43.04% Token: | blue|
Top 0th token. Logit: 24.22 Prob: 56.15% Token: | green|
Top 1th token. Logit: 23.95 Prob: 43.04% Token: | blue|
\end{lstlisting}
\end{promptbox}
\newpage
\subsection{Notable features}
Table \ref{tab:notable_features} shows notable feature found in the SAE of Gemma and LLaMA models and their impact on the average 4th round defection probability in the IPD.
\begin{table*}[!htbp]
\centering
\caption{Semantic interpretation and probability analysis for different models and features.}
\label{tab:semantic_probabilities}
\begin{tabular}{@{}llclcc@{}}
\toprule
\textbf{Interpretation} & \textbf{Model} & \textbf{Feature ID} & \textbf{Semanticity} & \(\displaystyle \mathbf{\left\langle P\left(\textrm{`blue'}|+\right)\right\rangle}\) & \(\displaystyle \mathbf{\left\langle P\left(\textrm{`blue'}|-\right)\right\rangle}\) \\ 
\midrule
Green          & Gemma-2b     & 1041       & Mono         & 0.23                                                              & 0.95  \\
Blue           & Gemma-2b     & 6556       & Mono         & 0.80                                                              & 0.04  \\
Change         & Gemma-2b     & 6879       & Poly         & 0.47                                                              & 0.84  \\
Sacrifice      & Gemma-2b     & 7155       & Poly         & 0.22                                                              & 0.69  \\
Trust          & Gemma-2b     & 7445       & Mono         & 0.50                                                              & 0.47  \\

Environment   & Gemma2-2b    & 6167       & Mono         & 0.34                                                              & 0.84  \\
Blue           & LLaMA3-IT-8b & 45097      & Mono         & 0.81                                                              & 0.27  \\

Environment   & LLaMA3-IT-8b & 15699      & Mono         & 0.42                                                              & 0.95  \\
Good/Bad Faith & LLaMA3-IT-8b & 30695      & Mono         & 0.75                                                              & 0.47  \\
\bottomrule
\end{tabular}
\label{tab:notable_features}
\end{table*}

We find that steering with `change' feature allows us to shift the last token probability distribution by $37$ percentage points. With steering, the probability of defection, averaged over all 64 histories changed from:
\begin{align*}
    \left\langle P\left(\textrm{` blue'}| +change\right)\right\rangle = 47\%,
\end{align*}
for the positive $w$ to
\begin{align*}
    \left\langle P\left(\textrm{` blue'}|-change\right)\right\rangle = 84\%,
\end{align*}
for the negative value of $w$.

Even though not for every history configuration the cooperation probability rises, we see that `change'-steered model shifts the generation towards cooperation. It seems that the removing the `change' feature effect from the residual stream (by subtracting it with the weight $w$) makes the agent's decisions much less cooperative.

\begin{figure}[!htbp]
     \centering
     \includegraphics[width=0.6\columnwidth]{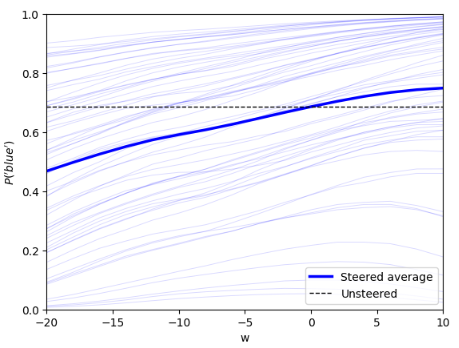}
     \caption{llama-3 steering with good/bad faith feature.}
     \label{fig:llama3-good/bad_faith_monotonic}
\end{figure}

Indeed, there are several features that drastically change the probability of defection. We list three notable features from Gemma-2b:
\begin{itemize}
    \item Steering towards \textbf{green} feature (1041) drastically changes the ` blue' token generation probability:
\begin{align*}
    \left\langle P\left(\textrm{` blue'}|-green\right)\right\rangle &= 95\%,\\    \left\langle P\left(\textrm{` blue'}|+green\right)\right\rangle &= 23\%
\end{align*}
this is sanity test result, as `green' is the other action option, corresponding to the cooperation. Biasing the model towards this direction, decreases the ` blue' token probability, since both of these probabilities should sum approximately to one:
$P(\textrm{` green'}) + P(\textrm{` blue'}) \sim 1$ 
\item Steering towards \textbf{blue} feature (6556) drastically changes the ` blue' token generation probability:
\begin{align*}
    \left\langle P\left(\textrm{` blue'}|-blue\right)\right\rangle &= 0.04\%,\\     \left\langle P\left(\textrm{` blue'}|+blue\right)\right\rangle &= 80\%
\end{align*}

\item The other, polysemantic feature (7155) is more interesting - it activates most strongly on tokens used for descriptions of \textbf{sacrifice, hope, victims}, but also on \textbf{sentences opposing violence}.
\begin{align*}
    \left\langle P\left(\textrm{` blue'}|-sacrifice\right)\right\rangle &= 69\%, \\ \left\langle P\left(\textrm{` blue'}|+sacrifice\right)\right\rangle &= 22\%
\end{align*}

When steered in the direction opposing `sacrifice' the model becomes much more vindictive. On the other hand, positive steering towards `sacrifice' results usually with cooperation.

\item A monosemantic LLaMA3-IT-8b feature \textbf{good/bad faith negotiation} has number of desirable qualities for the LLM Agent alignment 
\begin{align*}
    \left\langle P\left(\textrm{` blue'}|-\textrm{good/bad faith}\right)\right\rangle &= 47\%, \\ \left\langle P\left(\textrm{` blue'}|+\textrm{good/bad faith}\right)\right\rangle &= 75\%
\end{align*}

It is monosemantic, as evidenced by its top token activations and the activation density histogram (see Figure~\ref{positive_activations_histogram_good_faith} in the appendix). Its meaning is relevant to human moral values. It aligns with our intuition and changes the average defection probability by a substantial amount (28 percentage points). It exhibits an approximately monotonic relationship (see Figure~\ref{fig:llama3-good/bad_faith_monotonic}) with steering strength for each history combination. 
\end{itemize}

This shows that both the surface-level result of action like `green' and abstract ideas like ` debt' and the `denounce of violence' greatly affect the action choice.

\subsection{Activation dashboards}
As described in \citep{bricken2023towards}, the attention activation dashboards allow us to indirectly investigate the monosemanticity of a given feature. One can pass a dataset through a model and collect the activation values of a chosen feature. The distributions of such collected data are shown in the figures below.

\begin{figure*}[!htbp]
\vskip 0.2in
\begin{center}
    \centerline{
        \includegraphics[width=0.48\textwidth]{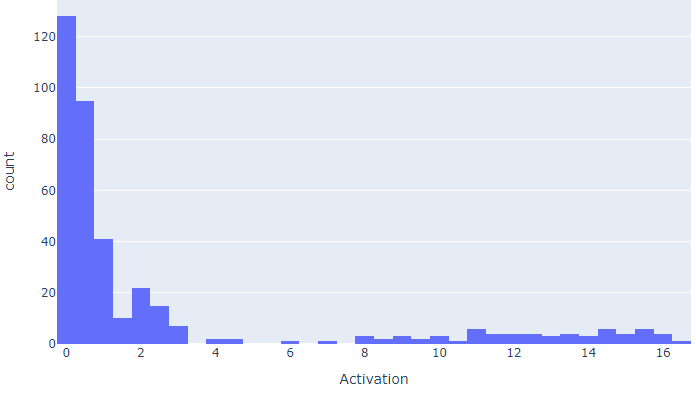}
        \hspace{0.02\textwidth} % Adjust spacing between the two figures
        \includegraphics[width=0.48\textwidth]{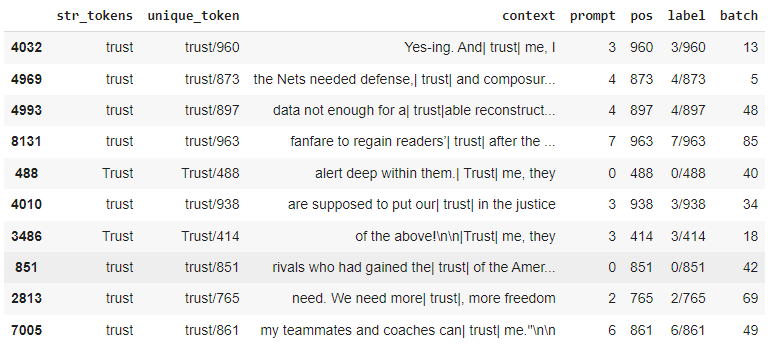}
    }
\end{center}
\vskip -0.2in
\caption{Activation density distribution - `trust' feature of Gamma-2b. Top 10 activations on the feature}
\label{fig:trust_histogram}
\end{figure*}

There are two general classes of the distributions:
\begin{itemize}
    \item flat tail distribution - the bin counts monotonically decrease with the strength of the activation. There are no text samples which disproportionately activate features. See e.g. `sacrifice' feature in Figure~\ref{fig:sacrifice_histogram}
    \item tail cluster - the bin counts have bimodal distribution with smaller mode in the large activation values. See e.g. `trust' in Figure~\ref{fig:trust_histogram}
\end{itemize}

\begin{figure*}[!htbp]
\vskip 0.2in
\begin{center}
    \centerline{
        \includegraphics[width=0.48\textwidth]{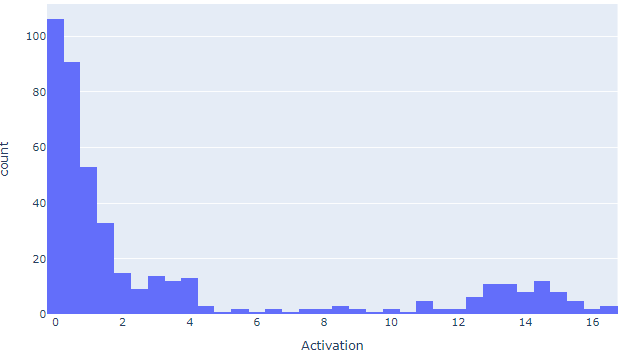}
        \hspace{0.02\textwidth} % Adjust spacing between the two figures
        \includegraphics[width=0.48\textwidth]{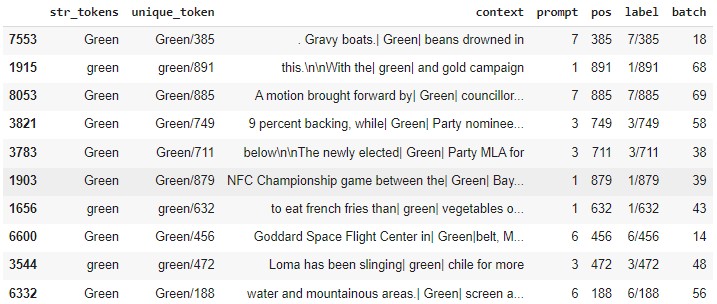}
    }
\end{center}
\vskip -0.2in
\caption{Activation density distribution - `Green' feature of Gamma-2b. Top 10 activations on the feature}
\label{fig:green_histogram}
\end{figure*}

\begin{figure*}[!htbp]
\vskip 0.2in
\begin{center}
    \centerline{
        \includegraphics[width=0.48\textwidth]{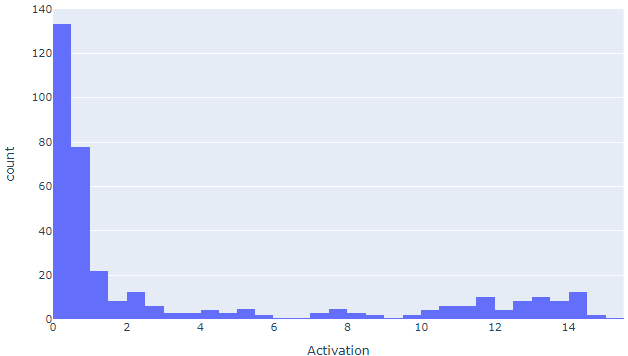}
        \hspace{0.02\textwidth} % Adjust spacing between the two figures
        \includegraphics[width=0.48\textwidth]{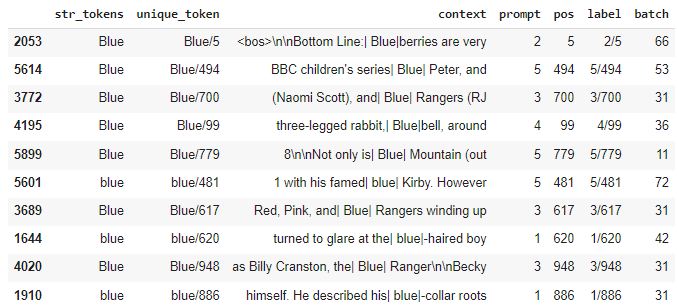}
    }
\end{center}
\vskip -0.2in
\caption{Activation density distribution - `blue' feature of Gamma-2b. Top 10 activations on the feature}
\label{fig:blue_histogram}
\end{figure*}

\begin{figure*}[!htbp]
\vskip 0.2in
\begin{center}
    \centerline{
        \includegraphics[width=0.48\textwidth]{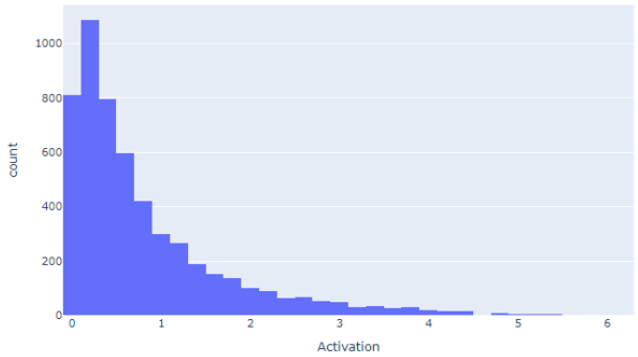}
        \hspace{0.02\textwidth} % Adjust spacing between the two figures
        \includegraphics[width=0.48\textwidth]{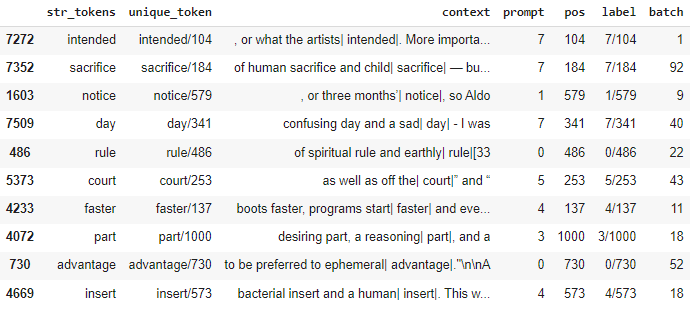}
    }
\end{center}
\vskip -0.2in
\caption{Activation density distribution - polysemantic feature `sacrifice' of Gamma-2b. Top 10 activations on the feature}
\label{fig:sacrifice_histogram}
\end{figure*}

\begin{figure*}[!htbp]
\vskip 0.2in
\begin{center}
    \centerline{
        \includegraphics[width=0.48\textwidth]{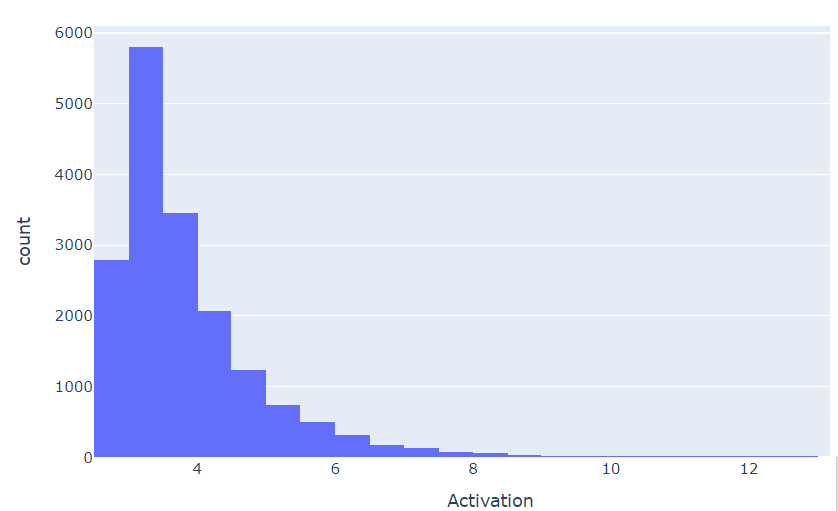}
        \hspace{0.02\textwidth} % Adjust spacing between the two figures
        \includegraphics[width=0.43\textwidth]{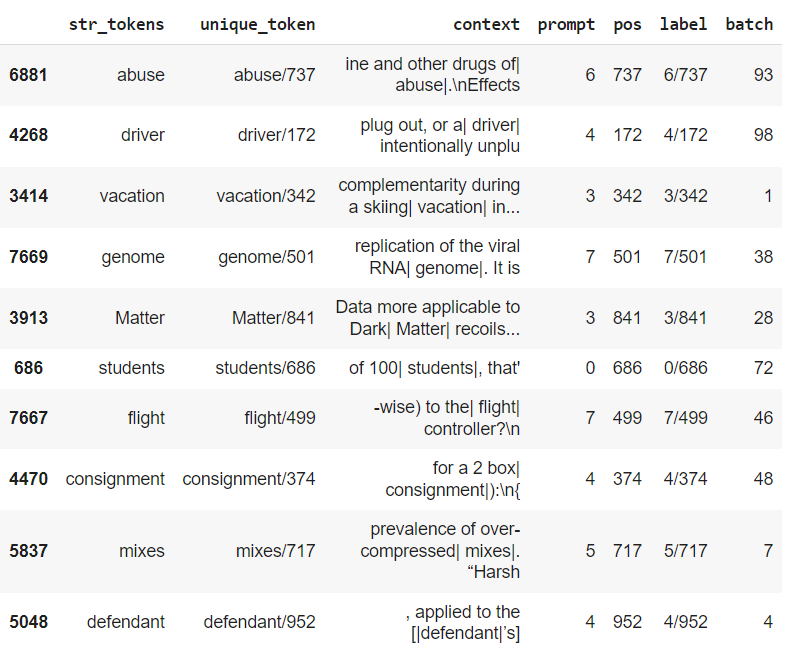}
    }
\end{center}
\vskip -0.2in
\caption{Activation density distribution - polysemantic feature `abuse' of LLaMA3-it-8b. Top 10 activations on the feature}
\label{fig:sacrifice_histogram}
\end{figure*}

\begin{figure*}[!htbp]
\vskip 0.2in
\begin{center}
    \centerline{
        \includegraphics[width=0.48\textwidth]{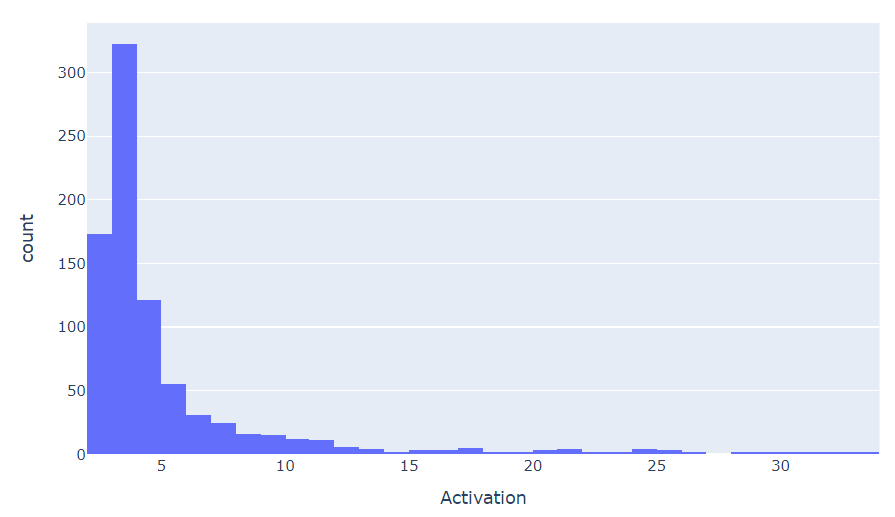}
        \hspace{0.02\textwidth} % Adjust spacing between the two figures
        \includegraphics[width=0.48\textwidth]{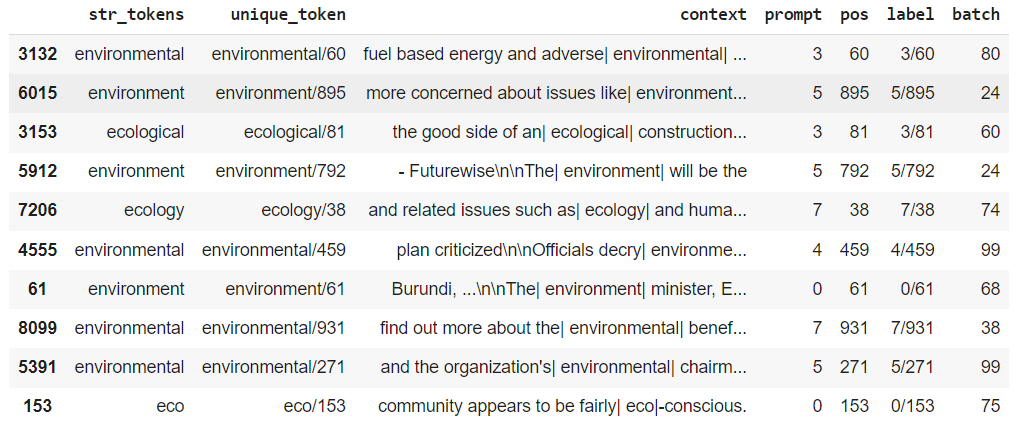}
    }
\end{center}
\vskip -0.2in
\caption{Activation density distribution - feature `environment' of Gemma2-2b. Top 10 activations on the feature}
\label{fig:sacrifice_histogram}
\end{figure*}

\begin{figure*}[!htbp]
\vskip 0.2in
\begin{center}
    \centerline{
        \includegraphics[width=0.48\textwidth]{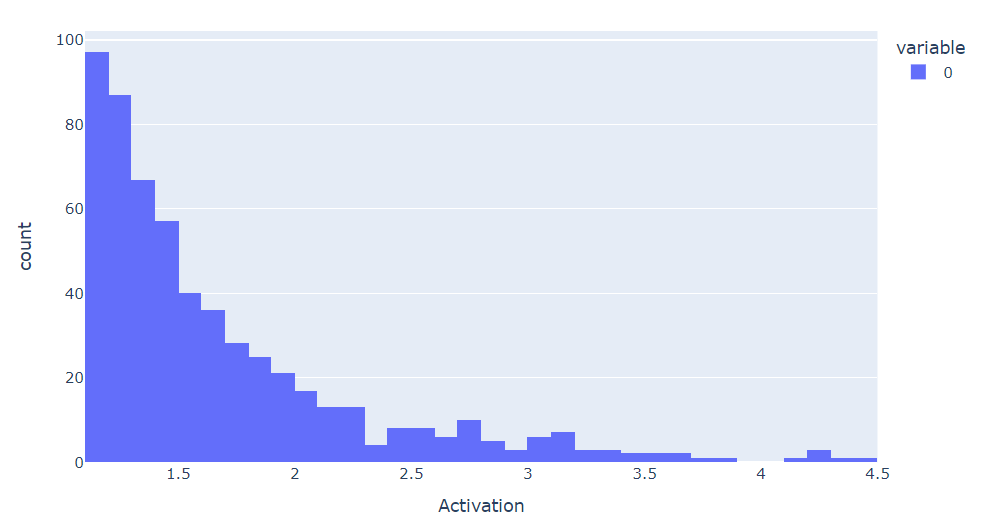}
        \hspace{0.02\textwidth} % Adjust spacing between the two figures
        \includegraphics[width=0.48\textwidth]{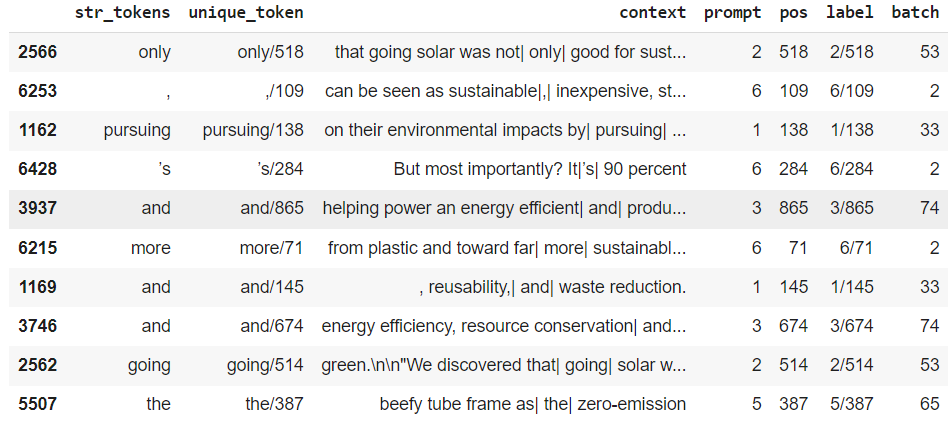}
    }
\end{center}
\vskip -0.2in
\caption{Activation density distribution - feature `good faith' of LLaMA3-it-8b Top 10 activations on the feature}
\label{positive_activations_histogram_good_faith}
\end{figure*}

\begin{figure*}[!htbp]
\vskip 0.2in
\begin{center}
    \centerline{
        \includegraphics[width=0.48\textwidth]{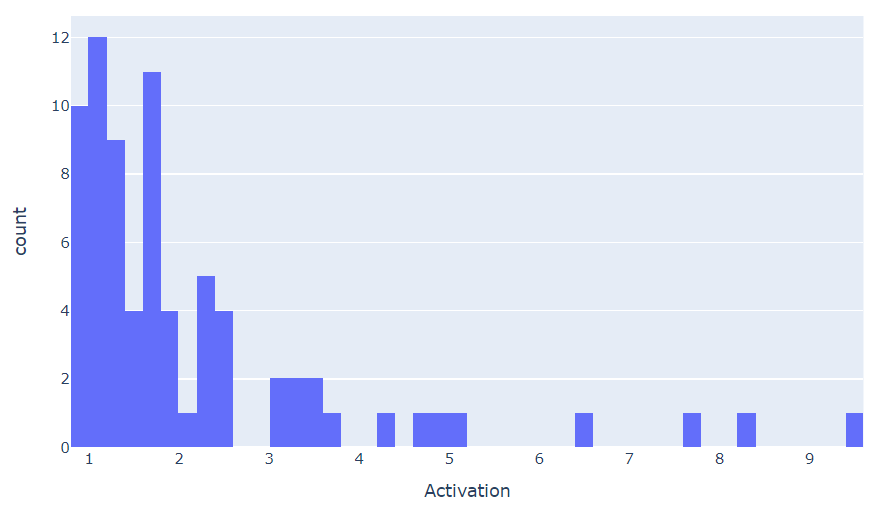}
        \hspace{0.02\textwidth} % Adjust spacing between the two figures
        \includegraphics[width=0.48\textwidth]{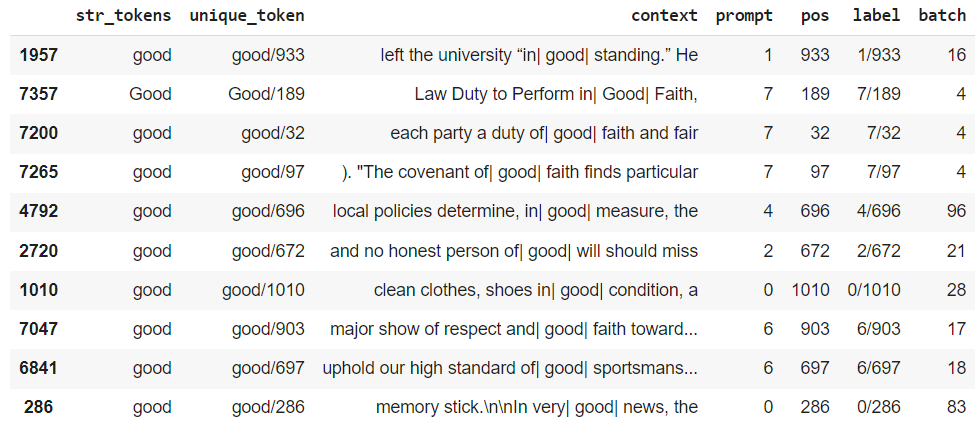}
    }
\end{center}
\vskip -0.2in
\caption{Activation density distribution - feature `good faith' of LLaMA3-it-8b. Top 10 activations on the feature}
\label{positive_activations_histogram_abuse}
\end{figure*}

\end{document}